\documentclass[letterpaper, 10 pt, conference]{ieeeconf}

\IEEEoverridecommandlockouts  
\usepackage{algorithm}
\usepackage{algorithmic}
\usepackage{amsmath}
\usepackage{amssymb}
\usepackage{booktabs}
\usepackage{cite}
\usepackage{float}
\usepackage{glossaries}
\usepackage{graphicx}
\usepackage{lipsum}
\usepackage{multirow}
\usepackage{siunitx}
\usepackage{xcolor}
\usepackage[hidelinks]{hyperref}

\DeclareMathOperator*{\argmax}{arg\,max}

\newacronym{fcn}{FCN}{Fully Convolutional Network}
\newacronym{ig}{IG}{Information Gain}
\newacronym{ik}{IK}{Inverse Kinematics}
\newacronym{ros}{ROS}{Robot Operating System}
\newacronym{tsdf}{TSDF}{Truncated Signed Distance Function}
\newacronym{vgn}{VGN}{Volumetric Grasping Network}

\title{\LARGE \bf
Closed-Loop Next-Best-View Planning for Target-Driven Grasping
}

\author{Michel Breyer, Lionel Ott, Roland Siegwart, and Jen Jen Chung%
\thanks{This work was supported in part by the Luxembourg National Research Fund (FNR) 12571953 and ABB Corporate Research.}%
\thanks{Autonomous Systems Lab, ETH Zurich, 8092 Zurich, Switzerland, \texttt{\{mbreyer, lioott, rsiegwart, chungj\}@ethz.ch.}}%
}

\begin{document}

\maketitle


\begin{abstract}
Picking a specific object from clutter is an essential component of many manipulation tasks. Partial observations often require the robot to collect additional views of the scene before attempting a grasp. This paper proposes a closed-loop next-best-view planner that drives exploration based on occluded object parts. By continuously predicting grasps from an up-to-date scene reconstruction, our policy can decide online to finalize a grasp execution or to adapt the robot's trajectory for further exploration. We show that our reactive approach decreases execution times without loss of grasp success rates compared to common camera placements and handles situations where the fixed baselines fail. Video and code are available at \url{https://github.com/ethz-asl/active_grasp}.
\end{abstract}


\section{Introduction}

\begin{figure}[t!]
  \centering
  \includegraphics[width=\columnwidth]{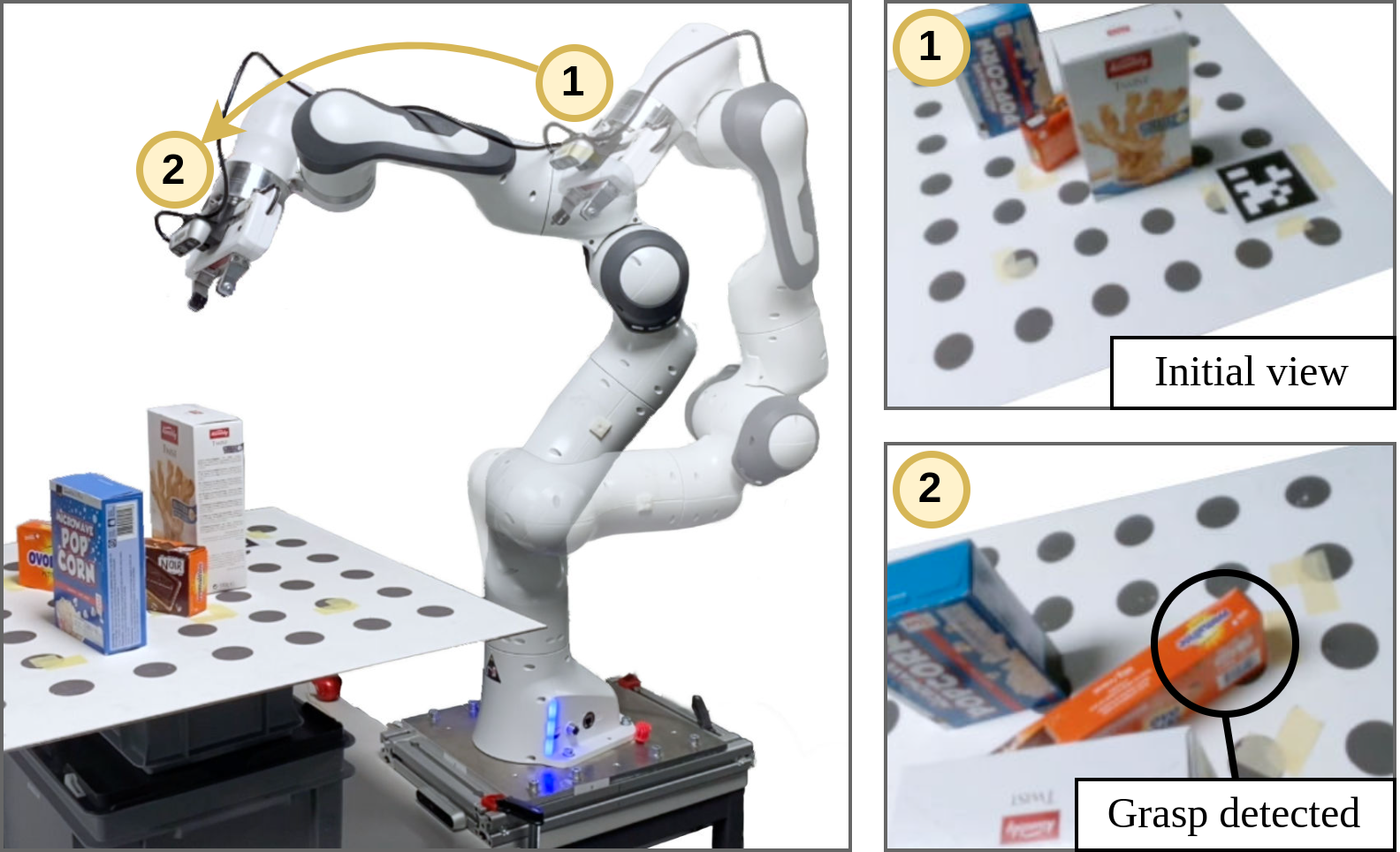}
  \caption{Detecting grasps on the orange packet using the wrist-mounted camera is hindered by occlusions from surrounding objects. By moving the sensor, our next-best-view grasp planner successfully discovers a grasp on the target item.}
  \label{fig:motivation}
\end{figure}

Planning grasps using on-board sensing is an essential skill for robots to intelligently interact with the unstructured environments outside of assembly lines. While tremendous progress has been made in predicting grasps from partial point cloud data~\cite{Mahler2017,Sundermeyer2021}, retrieving a target object from clutter remains challenging. One example is shown in Fig.~\ref{fig:motivation} where a manipulator with a wrist-mounted camera is tasked to grab the orange packet of cookies. The item is clearly visible from the initial view, but the graspable region is occluded by the surrounding objects. In order to complete the task, the robot has to move its sensor and actively explore the environment to discover a grasp on the target.

Previous works have proposed active perception systems to aid grasp synthesis. However, they often completely separate exploration and grasp detection~\cite{Bone2008,Kahn2015}, or require an initial grasp hypothesis to guide viewpoint selection~\cite{Arruda2016, Gualtieri2017, Morrison2019}. In this paper, we focus on designing a single policy that efficiently discovers stable grasp configurations on a partially occluded target object in clutter. 

Our method combines next-best-view planning~\cite{Isler2016} and real-time grasp detection~\cite{Breyer2020}. We continuously integrate new sensor measurements into a volumetric map of the scene, detect grasps, and compute the next viewpoint based on a count of voxels that are likely to belong to the target object. By re-planning at a rate of \SI{4}{\hertz}, our approach is able to quickly adjust the robot's trajectory based on the updated information, reducing overall execution times. In addition, a task-driven termination criterion is used to let our algorithm decide whether to execute a detected grasp at any point along the robot trajectory. However, to reduce failures, we only commit to a grasp configuration once it remains stable even after integrating new observations.

We validate our approach in simulation and on a real robotic setup and compare our results to several baselines that represent common camera placements for grasp detection. We found that for scenarios that favor top-down grasps, our approach achieves similar success rates compared to top-down camera placements while on average reducing the search time by around 40\%. In addition, we show that our method can adapt to more diverse situations, solving problems where the fixed baselines fail.

In summary, the contributions of this work are:

\begin{enumerate}
    \item an online next-best-view planner for discovering grasps on a target object in clutter,
    \item a task-driven termination criterion based on past grasp predictions,
    \item experimental results showing our approach handling diverse situations faster and more robustly compared to baselines.
\end{enumerate}

\section{Related Work}

Recent advances in object-agnostic grasp synthesis have been mostly driven by deep learning approaches that locate grasps directly in single depth image observations~\cite{Mahler2017,Sundermeyer2021}. However, the robustness of such systems can be severely impacted by the amount of available visual information. For example, Gualtieri et al.~\cite{Gualtieri2016} showed that fusing observations from many viewpoints significantly increased precision and recall compared to two static cameras. One way to tackle the challenge of partial observations is to train models to complete the occluded regions~\cite{Lundell2020,Jiang2021}. However, high levels of occlusion still motivate an approach that physically moves the sensor to gather more information. 

Active perception~\cite{Bajcsy2018} studies the problem of determining sensor placements that maximize the information collected from the environment and has been used in various applications such as mapping~\cite{Popovic2020,Schmid2020}, object search~\cite{Xiao2019,Novkovic2020}, pose estimation~\cite{Kim2016, Murali2022}, and 3D reconstruction~\cite{Kriegel2015, Daudelin2017}. A common approach is to select the ``best'' next view according to some \gls{ig} measure~\cite{Connolly1985}. However, choosing a metric can be challenging and is typically highly task specific~\cite{Chen2011}. In this work, we take inspiration from the rear side voxel \gls{ig} formulation proposed by Isler et al.~\cite{Isler2016} for volumetric reconstruction of objects by a mobile robot.

Active vision systems have also been used to aid grasp synthesis. Arruda et al.~\cite{Arruda2016} propose an \gls{ig} formulation that maximizes surface reconstruction close to the contact points between the object and a given grasp. Gualtieri and Platt~\cite{Gualtieri2017} directly use the output of their grasp detection pipeline to construct an object-class-specific database of informative viewpoints while Morrison et al.~\cite{Morrison2019} collect data online guided by the entropy in accumulated pixel-wise grasp predictions. Common among these works is that the next view is driven by an initial grasp detection.

In contrast, Kahn et al.~\cite{Kahn2015} plan sensor motion to discover grasp handles within occluded regions of the scene following a classic sense-plan-act approach. Chen et al.~\cite{Chen2020} train a reinforcement learning policy to increase the visibility of a target object in clutter and to stop exploration to plan grasps.

We combine ideas from both lines of work. Our next-best-view grasp planner explores occluded parts of the object while continuously generating grasp hypotheses, and we use the distribution of past predictions to fix a final configuration.

While most grasp planners are target-agnostic, some works have looked into picking a specific object, either through reinforcement learning~\cite{Jang2017, Fujita2020} or by using instance segmentation to match grasps with the target object~\cite{Murali2020}. We follow the latter approach and assume that a bounding box of the desired item is provided.


\section{Approach}

We consider the problem of moving a depth camera attached to the hand of a robotic arm in order to find a parallel-jaw grasp on a given target object. We make the common assumption that the calibration between the optical center of the camera and the end effector is known. In addition, since we are interested in picking a specific object, we assume that we start with a partial view and a 3D bounding box of the target.

\begin{figure}
  \centering
  \includegraphics[width=\columnwidth]{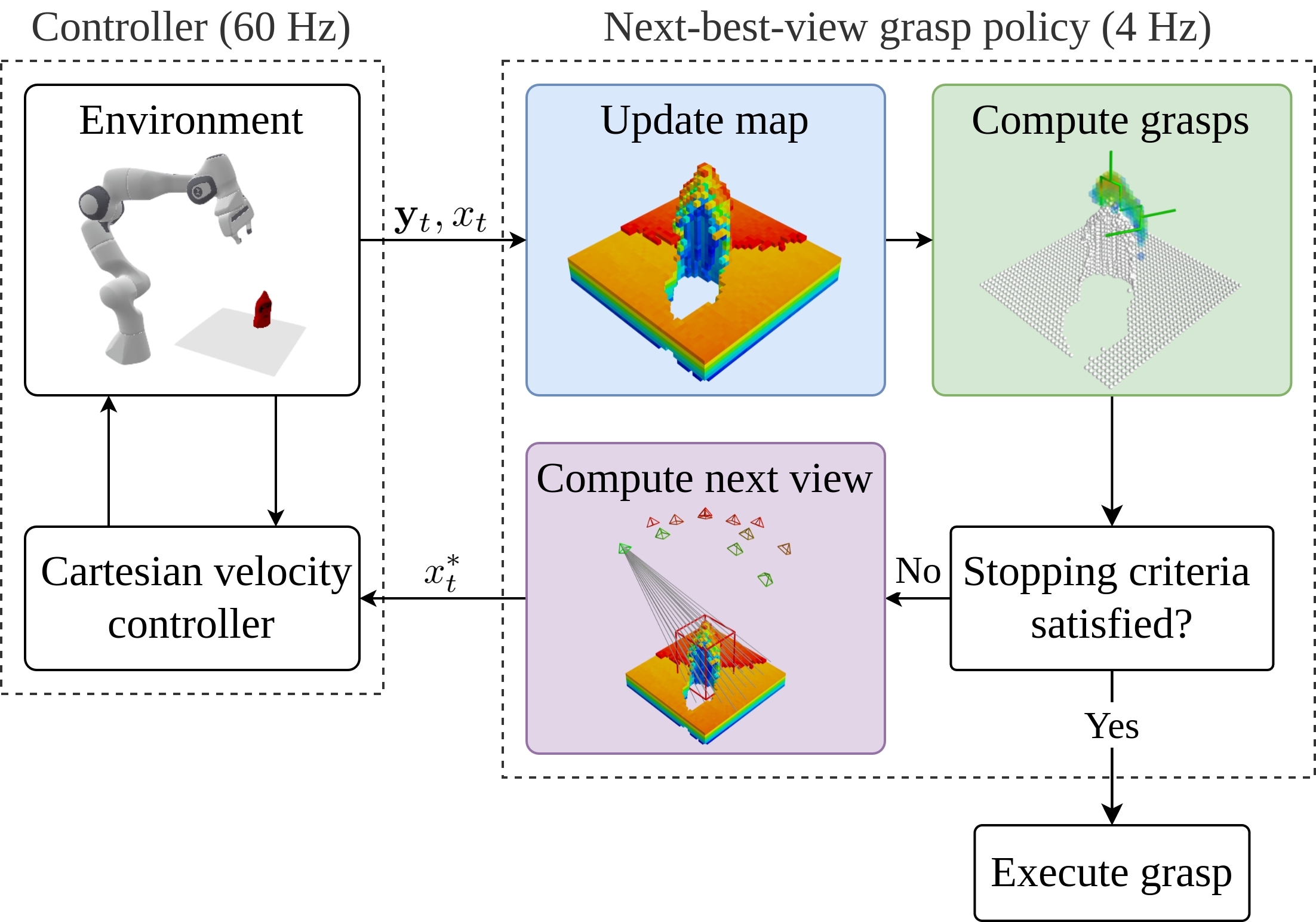}
  \caption{Overview of the framework. Our policy continuously integrates sensor measurements into a volumetric map of the scene, computes grasps, and re-plans informative views until a stable grasp is detected.}
  \label{fig:overview}
\end{figure}

Our goal is to find a policy which continuously processes the latest sensor measurements and either returns a stable grasp detection or an informative view towards which the robot should move. An overview of our system is shown in Fig.~\ref{fig:overview}. At every time step $t$, we integrate the current point cloud observation $y_t$ and camera pose $x_t$ into a \gls{tsdf}~\cite{Curless1996} reconstruction of the scene and compute voxel-wise grasp affordances ~\cite{Breyer2020}. Next, we compute the view $x^*_t$ from which we expect to observe occluded parts of the target item, which is then tracked by a Cartesian velocity controller. Since our primary goal is grasp success (rather than object reconstruction, etc.), we use a stopping criterion based on the history of grasp predictions to determine whether a suitable configuration has been found. By evaluating the policy at a fixed rate, the system can continuously move towards views based on expected information gain, but will execute a grasp whenever the stopping criterion is met. We first present the grasp detection and view planning components in Sections~\ref{sec:grasp-detection} and~\ref{sec:nbv}, before we combine them into a fully autonomous exploration and grasping system in Section~\ref{sec:controller}.

\subsection{Grasp Detection}\label{sec:grasp-detection}

We use the \gls{vgn} for grasp synthesis. \gls{vgn} is a fully convolutional network that maps a voxel grid $M$ to a grasp quality score $Q$, along with the associated orientation $R$ and opening width $W$ of a parallel-jaw grasp at every voxel. Given the output of the pre-trained network, we follow the steps of~\cite{Breyer2020} to construct a list of promising grasp configurations $\xi_i$. Since we are only interested in grasps on the target object, we filter out the hand poses for which the fingertips do not lie within the target's bounding box. In addition, to avoid unreachable configurations, we remove grasps for which no \gls{ik} solution can be found. Finally, we mark the configuration $\xi^*$ with the highest predicted grasp quality as the current best grasp.

\subsection{Next-Best-View Planner} \label{sec:nbv}

Let $x$ be a view from a set of potential sensor placements $\mathcal{X} \subset SE(3)$. We define $\mathcal{G}_x$ as the predicted information gained by observing the scene from $x$. The goal of the next-best-view planner is to find the view with the highest predicted \glsfirst{ig},
\begin{equation}
    x^* = \argmax_{x \in \mathcal{X}}{\mathcal{G}_x}.
\end{equation}

The choice of world representation, sensor placements and information gain are crucial for any active perception task. In the following subsections, we present the different components of our grasp-driven view planner.

\subsubsection{World Representation}

We use a \glsfirst{tsdf} to represent a cubic volume of size $l$. A \gls{tsdf} is a three-dimensional grid $\mathcal{M}$ of uniformly sized voxels storing the projected truncated signed distance to the nearest surface. We chose \glsplural{tsdf} due to their efficient incremental updates and noise averaging properties. In addition, we can directly share the map between the \gls{ig} computation and grasp detection, avoiding the need for multiple maps.

\subsubsection{View Generation}\label{sec:view-generation}

We distribute a set of view candidates $\mathcal{X}$ on a hemisphere placed on top of the target's bounding box as shown in the purple block in Fig.~\ref{fig:overview}. The radius of the sphere is chosen such that the distances between the camera locations and the bounding box are greater than the minimum depth distance of the sensor. We use \gls{ik} to discard views that are not reachable by the arm. While more sophisticated methods for selecting the viewpoints are possible, we believe that the evenly spaced views provide sufficient resolution for our purpose.

\subsubsection{Information Gain}

It was shown that surface reconstruction quality and grasp predictions are tightly coupled~\cite{Gualtieri2016}. In the same way, when grasping using the VGN framework, the completeness of the \gls{tsdf} reconstruction often has a large impact on grasp discovery and prediction accuracy. For this reason, we propose to use a variant of the rear side voxel \gls{ig} formulation from Isler et al.~\cite{Isler2016}.

Partially observed objects will cast shadows, i.e. voxels with negative distance values on the occluded side. We use ray casting to count the number of hidden object voxels that would be revealed by placing the camera at any particular viewpoint. More precisely, for each view candidate $x \in \mathcal{X}$, we generate a set of rays $\mathcal{R}_x$ cast from the optical center through each pixel of a virtual camera placed at $x$. Each ray $r$ traverses a set of voxels $\mathcal{M}_r \subset \mathcal{M}$ until it hits an observed object surface. The gain $\mathcal{G}_x$ is then given by,
\begin{equation}
    \mathcal{G}_x = \sum_{r \in \mathcal{R}_x} \sum_{m \in \mathcal{M}_r} \mathcal{I}(m),
\end{equation}
where $\mathcal{I}(m) = 1$ if the voxel $m$ is located within the target's bounding box and is storing a negative distance value, $\mathcal{I}(m) = 0$ otherwise.  

\subsection{Closed-loop Exploration and Grasping}
\label{sec:controller}

For improved efficiency, we want to continuously process incoming sensor data and react to the updated information. For this reason, we evaluate our policy at a fixed rate. At each update $t$, we first update the \gls{tsdf} map with the latest sensor measurements. Next, we evaluate \gls{vgn} to determine the best grasp candidate $\xi_t^*$ and compute the next best view $x_t^*$ and associated \gls{ig} $\mathcal{G}_{x_t^*}$ as described in the previous sections. We then decide whether to stop the policy execution or to continue exploring based on three stopping criteria.

First, we impose a time budget in the form of a maximum number of policy updates $T_\text{max}$. Second, we terminate if $\mathcal{G}_{x_t^*}$ is below a given threshold $\mathcal{G}_\text{min}$, since this means that we don't expect to gain a meaningful amount of additional information. This is designed to capture cases where no feasible grasp configurations can be detected even after fully exploring the target object. Third, we stop once \gls{vgn} generates a grasp configuration that remained stable over several frames.

The intuition behind the third condition is that grasp detections can be imprecise and vary with updated map information due to occlusions, especially close to the boundaries of observed space. We formalize this as a function of the moving average of the predicted grasp quality at the voxel corresponding to the best grasp's location over a window of size $T$. More precisely, given the last $T$ grasp quality tensors $Q_{t-T:t}$ predicted by \gls{vgn} and the voxel $m$ corresponding to $\xi_t^*$, we consider the grasp stable if,
\begin{equation}
    \left( \frac{1}{T} \sum_{t'=t-T}^{t} Q_{t'}(m) \right) > \epsilon_{\mu}.
\end{equation}

If one of the stopping criteria is satisfied, we abort the policy and, if found, return the best grasp configuration $\xi_t^*$ to be executed by a separate controller. Otherwise, we set the next best view $x_t^{*}$ as the target of a Cartesian velocity controller. The velocity controller runs in a separate control loop at a higher rate and generates velocity commands of fixed magnitude in the direction of the most informative viewpoint while ensuring that the camera maintains a minimum distance to the target object.


\section{Evaluation}

The goal of our experiments is to answer the following two questions: (a) how does our active exploration approach compare in terms of efficiency and grasp success rates to common camera placement alternatives, and (b) how robust is our adaptive system against different object locations and amounts of occlusions compared to fixed baselines?

\subsection{Experimental Setup}

We evaluate our approach by attempting to grasp a target object in different scenarios (shown in Fig.~\ref{fig:scenarios}). Our test environment consists of a 7-DoF Panda arm with a RealSense D435 rigidly attached to the end effector. For each scenario, we place the target object at a predefined location, surround it with distractor objects, and move the robot arm to a fixed initial configuration. Note that small perturbations to the initial configuration and object locations lead to some variance over multiple runs for a given scenario. 

In addition to the real-world setup, we build a simulation environment in PyBullet~\cite{Coumans2016} that mimics the real system. The simulation allows us to run quantitative experiments over a large number of randomly generated scenes following the ``packed'' protocol from~\cite{Breyer2020}. For each scene, we render a segmentation mask of the initial view and choose the object with the smallest amount of visible pixels as target. Bounding boxes of the target are provided by the physics simulator.

\begin{table}
    \centering
    \caption{Parameters used in our experiments.}
    \begin{tabular}{l l l}
        \toprule
        \gls{tsdf} size & $l$ & $\SI{0.3}{\m}$ \\
        \gls{tsdf} voxel count per side & $N$ & $40$ \\
        Number of view candidates & $|\mathcal{X}|$ & $16$ \\
        Policy rate & & \SI{4}{\hertz} \\
        Maximum number of views & $T_\text{max}$ & $80$ \\
        Minimum \gls{ig} & $\mathcal{G}_\text{min}$ & 10 \\
        Grasp quality moving average window size & $T$ & $12$ \\
        Grasp quality threshold & $\epsilon_\mu$ & $0.9$ \\
        Linear velocity & & $\SI{5}{\cm / \s}$ \\
        \bottomrule
  \end{tabular}
  \label{table:parameters}
\end{table}

Our policy is implemented in Python using ROS for interfacing the hardware. The extrinsic parameters of the depth sensor are calibrated using the toolbox from Furrer et al.~\cite{Furrer2018}. We use the \gls{tsdf} implementation from Open3D~\cite{Zhou2018}, TRAC-IK~\cite{Beeson2015} for \gls{ik} computations, and Numba~\cite{Lam2015} to accelerate ray casting. The parameters of our approach are listed in Table~\ref{table:parameters}. All experiments were run on a computer equipped with an Intel Core i7-8700K and a GeForce GTX 1080 Ti.

To evaluate grasps returned by our policy, we first plan a trajectory to the grasp pose using MoveIt. To avoid collisions with other objects in the environment, we extract a point cloud representation of the \gls{tsdf}, compute clusters, and generate convex hulls for each segment which are added as collision objects to the MoveIt planning scene. Next, we close the gripper with constant force and count the grasp as a success if the target object was successfully lifted by \SI{10}{\cm}. Occasionally, MoveIt fails to find a path to the grasp returned by our policy. Since online collision-checking is out of the scope of this work, we remove these trials from our experiments, though we include a more detailed discussion about this problem in Section~\ref{sec:discussion}.

\subsection{Evaluation Metrics}

We evaluate the performance with the following metrics:

\begin{itemize}
    \item Success Rate (\textbf{SR}). Ratio of runs where the target was successfully grasped.
    \item Failure Rate (\textbf{FR}). Ratio of runs where a grasp was detected, but failed during execution.
    \item Aborted Rate (\textbf{AR}). Ratio of runs where no grasp on the target object was found.
    \item Mean number of \textbf{views}. Number of policy updates.
    \item \textbf{Search time}. Time elapsed between receiving a bounding box and returning a grasp configuration.
    \item \textbf{Total time}. Includes the time to execute a detected grasp in addition to the search time.
\end{itemize}

\subsection{Baselines}

We compare our method against three baseline policies that correspond to common camera placement strategies.

\begin{itemize}
    \item \emph{initial-view}: Detect grasps using a single image captured from the initial view and execute the best grasp.
    \item \emph{top-view}: Move the camera to the top of the view sphere described in Section~\ref{sec:view-generation} and detect grasps from a single top-down image. This is typically an effective strategy for table-top manipulation~\cite{Lundell2020}.
    \item \emph{top-trajectory}: Same as \emph{top-view} but integrate images along the trajectory to the top view.
\end{itemize}

All baselines use the same controller described in Section~\ref{sec:controller} to generate robot motion.

\subsection{Simulation Experiments}

\begin{table}[t!]
    \setlength{\tabcolsep}{3pt}
    \centering
    \caption{Results from the simulation experiments.}
    \begin{tabular}{l r r r r r r}
        \toprule
        \textbf{Policy} & \textbf{SR} & \textbf{FR} & \textbf{AR} & \textbf{Views}  & \textbf{Search time (s)} & \textbf{Total time (s)}\\
        \midrule
        \textit{initial-view} & $79$ & $3$ & $18$ & $1 \pm 0$ & $1.4 \pm 0.1$ & $16.2 \pm 2.2$ \\
        \textit{top-view} & $90$ & $3$ & $7$ & $1 \pm 0$ & $10.0 \pm 1.4$ & $22.5 \pm 1.5$ \\
        \textit{top-trajectory} & $91$ & $2$ & $7$ & $34 \pm 5$ & $9.5 \pm 1.3$ & $22.5 \pm 2.3$ \\
        \textit{nbv-grasp} & $89$ & $4$ & $7$ & $18 \pm 12$ & $5.9 \pm 3.1$ & $19.5 \pm 3.4$ \\
        \bottomrule
  \end{tabular}
  \label{table:results-sim}
\end{table}

\begin{figure}[b]
  \centering
  \includegraphics[trim=0 10 0 0, clip,width=\columnwidth]{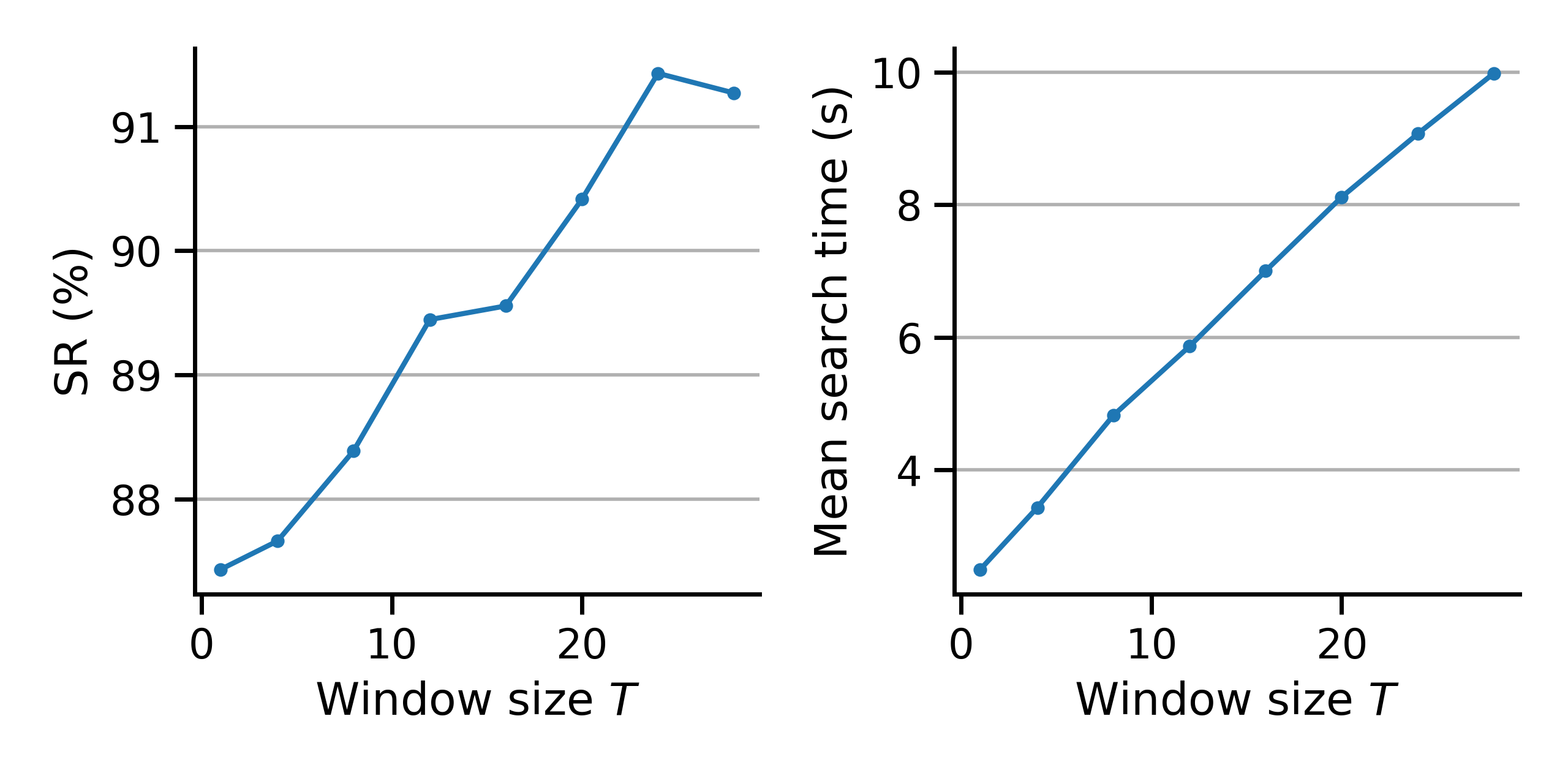}
  \caption{Success rate and search time vs window size $T$. Larger values of $T$ force the policy to explore longer, leading to higher success rates at the cost of longer search times.}
  \label{fig:effect_window_size}
\end{figure}

\begin{figure*}[t]
  \centering
  \includegraphics[width=\textwidth]{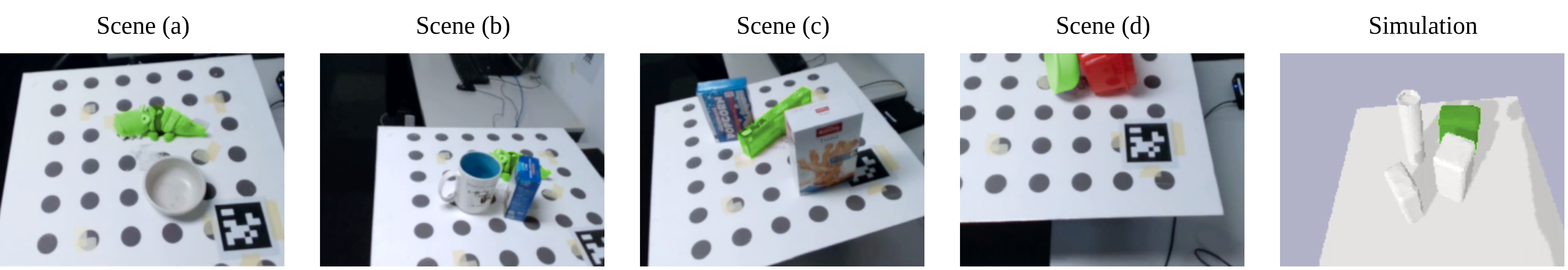}
  \caption{Images taken from the initial camera view for each test scene with the target object colored in green. Note that the RGB image has a smaller field of view compared to the depth sensor of the RealSense.}
  \label{fig:scenarios}
\end{figure*}

\begin{table*}[t]
    \centering
    \caption{Results from the real world case studies.}
    \begin{tabular}{l l r r r r r r}
        \toprule
        \textbf{Scene} & \textbf{Policy} & \textbf{SR} & \textbf{FR} & \textbf{AR} & \textbf{Views}  & \textbf{Search time (s)} & \textbf{Total time (s)}\\
        \midrule
        \multirow{4}{*}{(a)} & \textit{initial-view} & $4/4$ & $0/4$ & $0/4$ & $1 \pm 0$ & $1.9 \pm 0.7$ & $16.5 \pm 0.7$ \\
                             & \textit{top-view} & $4/4$ & $0/4$ & $0/4$ & $1 \pm 0$ & $9.8 \pm 0.6$ & $23.6 \pm 1.1$ \\
                             & \textit{top-trajectory} & $4/4$ & $0/4$ & $0/4$ & $33 \pm 1$ & $9.8 \pm 0.6$ & $23.2 \pm 1.0$ \\
                             & \textit{nbv-grasp} & $4/4$ & $0/4$ & $0/4$ & $18 \pm 7$ & $6.3 \pm 2.4$ & $20.9 \pm 2.6$ \\
        \hline
        \multirow{4}{*}{(b)} & \textit{initial-view} & $1/4$ & $1/4$ & $2/4$ & $1 \pm 0$ & $1.9 \pm 0.7$ & $17.1 \pm 0.4$ \\
                             & \textit{top-view} & $4/4$ & $0/4$ & $0/4$ & $1 \pm 0$ & $13.2 \pm 0.8$ & $26.8 \pm 0.9$ \\
                             & \textit{top-trajectory} & $4/4$ & $0/4$ & $0/4$ & $46 \pm 2$ & $13.2 \pm 0.9$ & $27.4 \pm 1.1$ \\
                             & \textit{nbv-grasp} & $4/4$ & $0/4$ & $0/4$ & $43 \pm 5$ & $12.6 \pm 0.9$ & $26.4 \pm 0.2$ \\
        \hline
        \multirow{4}{*}{(c)} & \textit{initial-view} & $1/4$ & $0/4$ & $3/4$ & $1 \pm 0$ & $1.9 \pm 0.8$ & $17.5$ \\
                             & \textit{top-view} & $4/4$ & $0/4$ & $0/4$ & $1 \pm 0$ & $13.5 \pm 0.8$ & $24.8 \pm 0.9$ \\
                             & \textit{top-trajectory} & $4/4$ & $0/4$ & $0/4$ & $48 \pm 1$ & $13.5 \pm 0.8$ & $25.0 \pm 0.9$ \\
                             & \textit{nbv-grasp} & $4/4$ & $0/4$ & $0/4$ & $30 \pm 15$ & $9.6 \pm 3.1$ & $23.1 \pm 2.2$ \\
        \hline
        \multirow{4}{*}{(d)} & \textit{initial-view} & $1/4$ & $0/4$ & $3/4$ & $1 \pm 0$ & $1.9 \pm 0.8$ & $34.5$ \\
                             & \textit{top-view} & $1/4$ & $0/4$ & $3/4$ & $1 \pm 0$ & $11.0 \pm 0.7$ & $23.8$ \\
                             & \textit{top-trajectory} & $2/4$ & $0/4$ & $2/4$ & $38 \pm 0$ & $11.1 \pm 0.7$ & $24.5 \pm 0.5$ \\
                             & \textit{nbv-grasp} & $4/4$ & $0/4$ & $0/4$ & $27 \pm 4$ & $8.5 \pm 1.7$ & $25.7 \pm 5.3$ \\

        \bottomrule
  \end{tabular}
  \label{table:results-hw}
\end{table*}

Table~\ref{table:results-sim} reports the results from running 400 trials with each policy. We observe that the \emph{initial-view} policy has the shortest total time since it does not explore the scene at all. However, this comes at the cost of a lower success rate as strong occlusions lead to a higher number of aborted runs. \emph{Top-view} and \emph{top-trajectory} resulted in similarly high success rates, even though \emph{top-trajectory} integrates more information about the scene along its trajectory. For the generated scenes, the target can typically be picked from the top, making the top view an effective strategy to discover grasps. Finally, the proposed \emph{nbv-grasp} policy has comparable grasping performance, but requires on average only little more than half the time to discover a successful grasp. The higher standard deviation in search time indicates that our policy adapts to the complexity of the scene, efficiently exploring and stopping once a stable grasp has been detected.

To investigate the influence of the stopping criterion, Fig.~\ref{fig:effect_window_size} shows SRs and mean search times for different values of the window size~$T$. We observe that higher values of $T$ result in higher success rates, as the policy is forced to explore longer, thus improving the accuracy of grasp detections in challenging cases. However, this also leads to increased mean search times.

\subsection{Real-world Experiments}

Table~\ref{table:results-hw} shows the results from four grasping trials for each of the four real-world scenarios shown in Fig.~\ref{fig:scenarios}. For scene~(a), where the object and initial camera are favorably placed, all policies succeed in retrieving the target. However, we can see that the \emph{nbv-grasp} policy operates more efficiently compared to the top-view baselines due to the early stopping criterion. Scenes (b) and (c) are designed similarly, with the target being visible, yet heavily occluded from the initial view. This is reflected in the degraded performance of the \emph{initial-view} policy, which fails in three out of the four attempts. Since the target can be grasped from the top, \emph{top-view} and \emph{top-trajectory} successfully handle these cases. The \emph{nbv-grasp} policy also succeeds in all cases, following a similar trajectory as the top-view baselines, as can be seen from the average search times and the accompanying video. Finally, the bowl placed on its side in scene (d) poses a challenge even for the top camera placement. Only the \emph{nbv-grasp} policy consistently explores the side views of the scene and detects a grasp on the rim, highlighting the ability of our approach to adapt to scenes of various complexity.

\subsection{Computation Times}

\begin{figure}[t]
    \centering
    \includegraphics[width=\columnwidth]{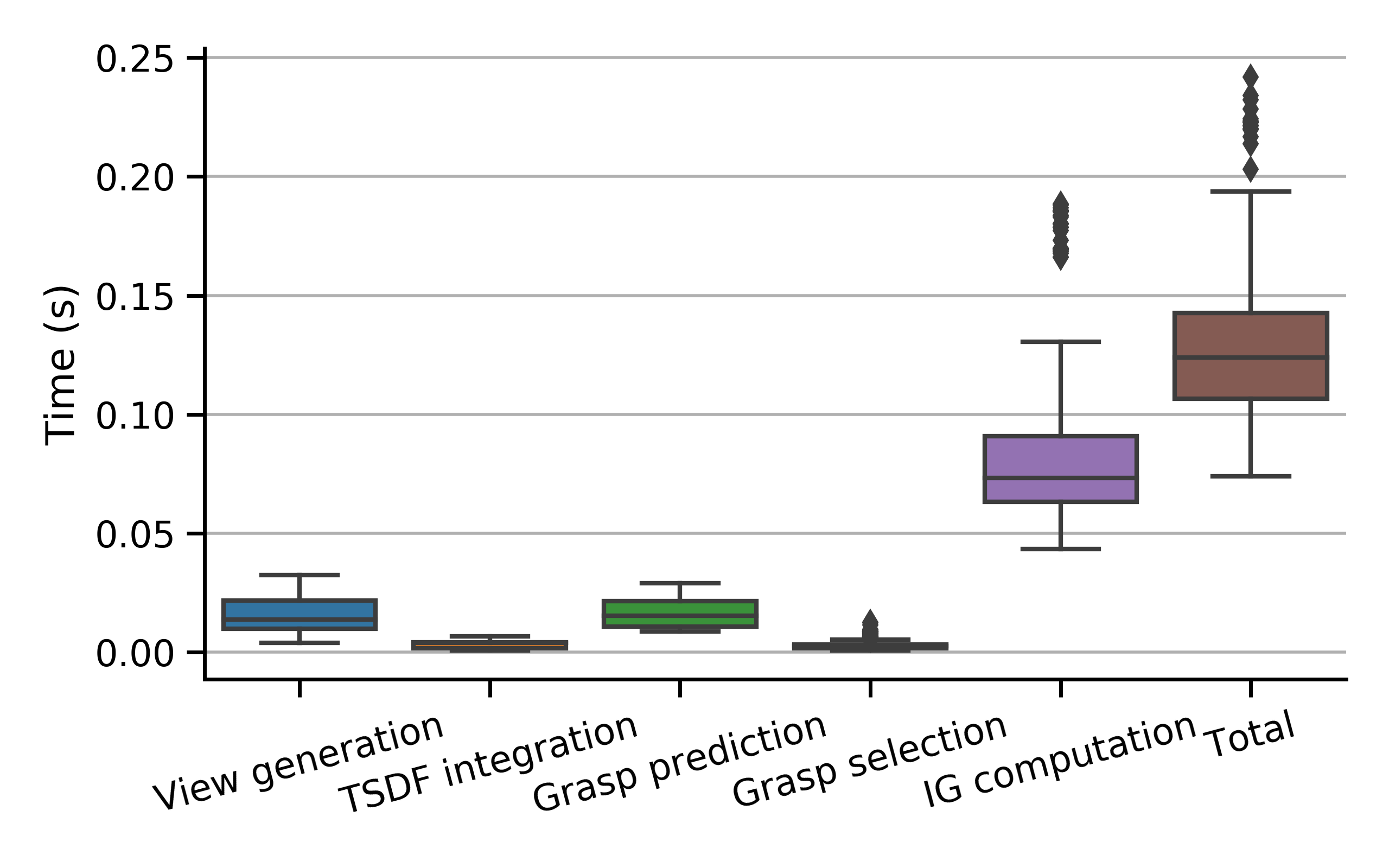}
    \caption{Computation times for one update of our \emph{nbv-grasp} policy measured over 40 simulation runs.}
    \label{fig:computation_times}
\end{figure}

Fig.~\ref{fig:computation_times} shows the computation times for one update of our \emph{nbv-grasp} policy measured over 40 simulation runs. We observe that the \gls{ig} computation accounts for the largest portion of total time. This could be improved by optimizing the ray casting implementation, however, the approach is still efficient enough to run online at 4 Hz.


\section{Discussion} \label{sec:discussion} 

The goal of this work is to enable a robot to efficiently retrieve a target object from clutter. We showed that our next-best-view policy often adapts a top-down strategy which is effective in many cases, but also generalizes to situations where the predefined camera locations fail. However, there remain several limitations leaving open directions for future research.

First, our policy considers the kinematic feasibility, but ignores collisions between the arm and the scene when computing grasp and view candidates. This can occasionally lead to situations where a grasp returned by our policy cannot be reached by the robot. This could be tackled with efficient collision checking, either based on geometric primitives~\cite{Pan2012} or learned collision functions~\cite{Danielczuk2021}.

Second, we assume the availability of bounding boxes for matching grasps to the target object. In a more integrated system, this could be replaced by object detection~\cite{Tekin2018} or instance segmentation~\cite{Xie2019}. It would also be interesting to include both object and grasp detection into a single active perception formulation.

Finally, we focused our study on the case where the robot can pick the target without additional manipulation. However, in dense clutter grasps on the target can be blocked by other objects, requiring multiple interactions, e.g. removing the blocking objects~\cite{Danielczuk2019, Murali2022}.


\section{Conclusion}

In this paper, we presented a next-best-view grasp planner that efficiently searches for stable grasp configurations on a partially occluded target object. Our information gain-based approach explores occluded parts of the object, adapting to the complexity of the scene. The presented stopping criteria help to balance exploration and exploitation by terminating the search once a stable grasp candidate has been found while reducing the number of false positive detections by enforcing a constraint on the grasp prediction history. We showed the increased robustness and efficiency of our approach compared to common fixed camera placement alternatives in a series of simulated and real world experiments.

In future work, we plan to investigate how to extend our formulation to more complex rearrangement tasks.

\bibliographystyle{IEEEtran}
\bibliography{IEEEabrv,references}

\end{document}